\documentclass[10pt,twocolumn,letterpaper]{article}

\usepackage{cvpr}              

\usepackage{graphicx}
\usepackage{amsmath}
\usepackage{amssymb}
\usepackage{booktabs}
\usepackage{algpseudocode}
\usepackage[accsupp]{axessibility}
\DeclareUnicodeCharacter{2212}{-}

\usepackage[pagebackref,breaklinks,colorlinks]{hyperref}

\usepackage[capitalize]{cleveref}
\crefname{section}{Sec.}{Secs.}
\Crefname{section}{Section}{Sections}
\Crefname{table}{Table}{Tables}
\crefname{table}{Tab.}{Tabs.}

\def\cvprPaperID{19} 
\def\confName{CVPR}
\def\confYear{2023}

\begin{document}

\title{Weakly Supervised Visual Question Answer Generation}

\author{Charani Alampalle\\
AlphaICs\\
Bengaluru, India\\
{\tt\small nagasai.charani@alphaics.com}
\and
Shamanthak Hegde\\
KLE Technological University\\
Hubballi, India\\
{\tt\small 01fe19bcs233@kletech.ac.in}
\and
Soumya Jahagirdar\\
CVIT, IIIT Hyderabad\\
Hyderabad, India\\
{\tt\small soumya.jahagirdar@research.iiit.ac.in}
\and
Shankar Gangisetty\\
IIIT Hyderabad\\
Hyderabad, India\\
{\tt\small shankar.gangisetty@ihub-data.iiit.ac.in}
}
\maketitle

\begin{abstract}
   Growing interest in conversational agents promote two-way human-computer communications involving asking and answering visual questions have become an active area of research in AI. Thus,  generation of visual question-answer pair(s) becomes an important and challenging task. To address this issue, we propose a weakly-supervised visual question answer generation method that  generates a relevant question-answer pairs for a given input image and associated caption. 
    Most of the prior works are supervised and depend on the annotated question-answer datasets. In our work, we present a weakly supervised method that synthetically generates question-answer pairs procedurally from visual information and captions. The proposed method initially extracts list of answer words, then does nearest question generation that uses the caption and answer word to generate synthetic question. Next, the relevant question generator converts the nearest question to relevant language question by dependency parsing and in-order tree traversal, finally, fine-tune a ViLBERT model with the question-answer pair(s) generated at end. We perform an exhaustive experimental analysis on VQA dataset and see that our model significantly outperform SOTA methods on BLEU scores. We also show the results wrt baseline models and ablation study.
\end{abstract}

\section{Introduction}
\label{sec:intro}

Conversational agents that can communicate with a human have been an active area of research and becoming popular due to artificial intelligence (AI). In conversational agent communication, visual question answering (VQA) is not only the desired characteristic where the agent answers a natural language question about the image, but also an intelligent agent should have the ability to ask a meaningful and relevant question with respect to its current visual perception. Keeping these points in mind, we can say that the best way to establish communication between humans and machines is by making machines how to ask meaningful and relevant question and at the same time, answer it properly when asked. Recently more focus is given to understanding the scene, asking and answering the questions from images and videos.

\begin{figure}
  \centering
  \includegraphics[width= 1\linewidth]{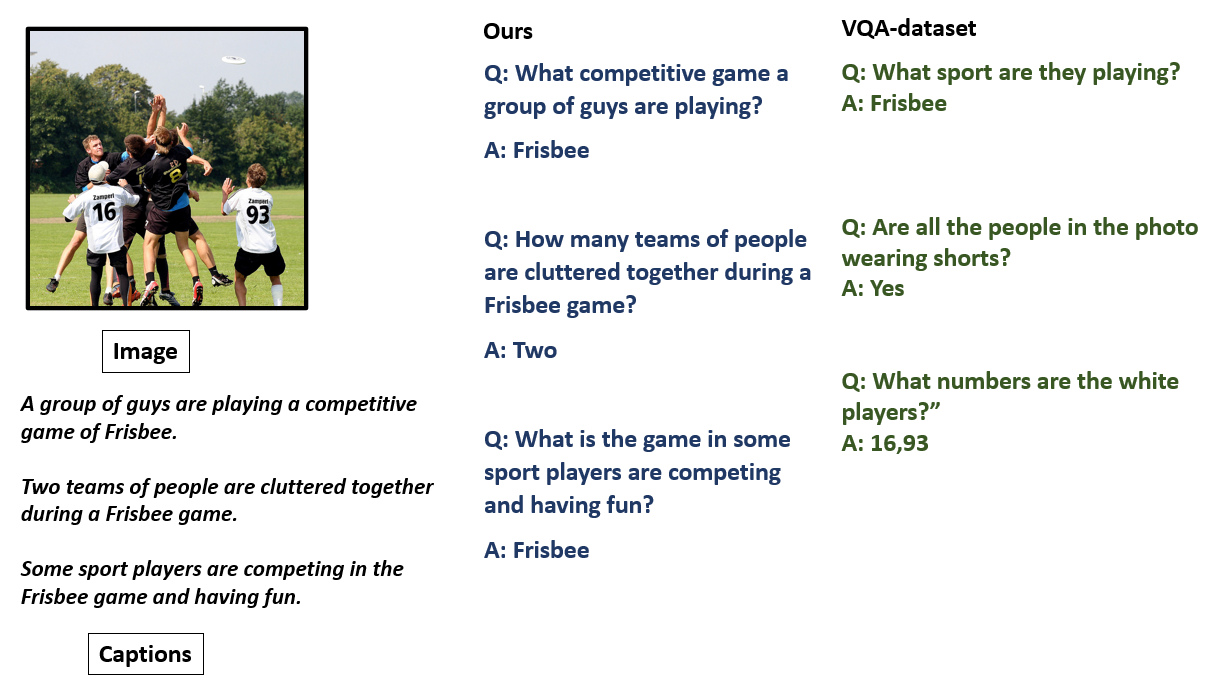}
  \caption{Our proposed weakly supervised VQA generation method gives importance to visual properties relevant to the given input image in order to generate specific question-answer pairs. Our results are automatically generated compared to VQA dataset~\cite{1} i.e., manually generated  question-answer pairs.}
  \label{fig:teaser}
\end{figure}

Works in various domains are mainly focusing on VQA~\cite{1} and visual question generation~\cite{5,6}. Given an image, generating meaningful questions, also known as visual question generation (VQG) is an essential component of a conversational agent~\cite{10} along with answering the questions. Visual question answer generation (VQAG) is a precursor of visual dialogue systems and might help in building large-scale VQA datasets automatically with less human effort. All the previous works~\cite{1,40,41} on question answer pair generation are dependent largely on datasets which was manual and tedious task to train. With less human effort to build VQA datasets needed for training, we build a VQAG model that can generate meaningful question-answer pairs for a given image and associated captions. Works in the direction of question-answer generation for image, given the caption focuses on generating generic questions or category based answers~\cite{8,9}. Unlike these we focus on category based questions (As categories we consider six types of question words like ``\textit{how many}”, ``\textit{who}”, ``\textit{what}”, ``\textit{which}”, ``\textit{how much}”, and ``\textit{where}") which help the conversational agents to communicate properly in understanding the context. Good question-answer pair is the one that has a tightly focused purpose and must be relevant to the image content. In this work, we fill-up the gap prevailing in the literature by introducing a method to solve the problem of generating question-answer pairs for the given image and associated captions with question being categorised. We refer our work as VQAG. 

\begin{figure}[h]
  \centering
  \includegraphics[width=\linewidth]{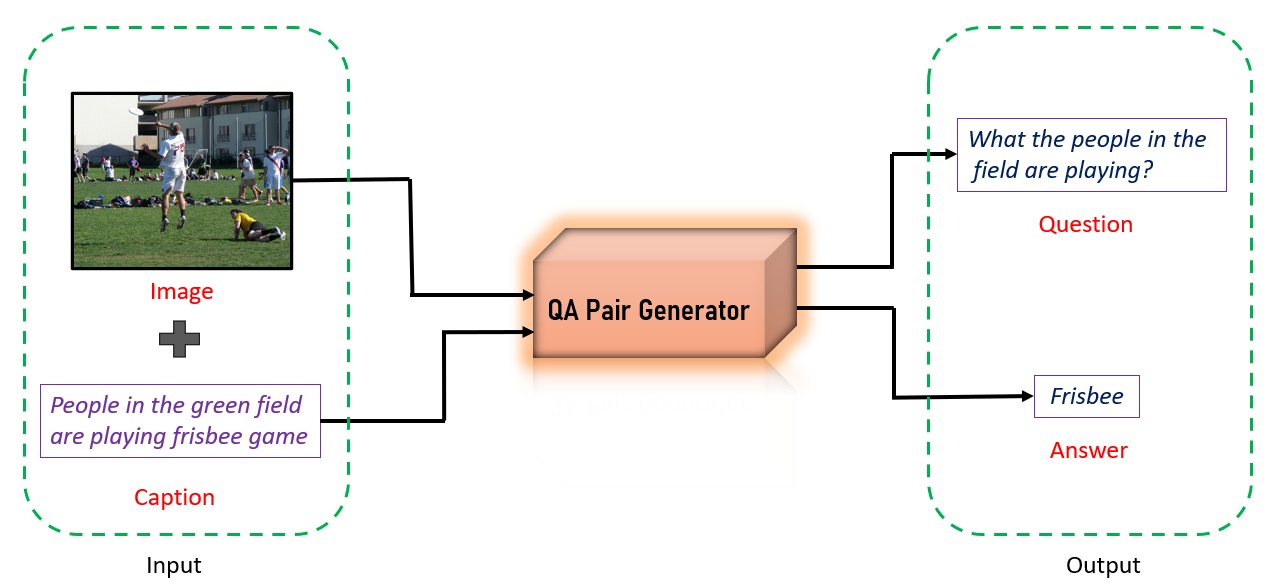}
  \caption{VQAG. We introduce a method of Visual Question Answer Generation. Given an image and its caption,  our goal is to generate a meaningful natural language question whose answer is one of the objects from image.}
  \label{fig:imvqg}
\end{figure}

Without depending on ground-truth QA pair(s) of images, the goal of VQAG is given a natural image and associated captions, list of objects are extracted from the image, generate a natural language question using caption, whose answer is one of the list of objects being identified. As we are not depending on ground-truth QA pair(s) and using captions instead we name our work as Weakly supervised Visual Question Answer Generation. As shown in Fig~\ref{fig:imvqg} given an image and associated caption(s), our aim is to generate a relevant language question such that the answer is in the list of the objects identified from the image. Here list of objects is [Fisbee, Person]. The problem is challenging as it requires in-depth semantic interpretation of the caption and from the image the visual content needs to generate meaningful and relevant questions. Question-answer pair generation has been a well-explored area in the language community~\cite{19}, vision and language community~\cite{8,9}. However, vision and language works often ignore the important visual information or objects appearing in the image, and only restrict themselves to the overall visual content while generating question and answer pairs. 

It should be noted that objects in the image helps not only in asking semantically meaningful and relevant questions connecting visual and textual content, but also helps to avoid generic questions as well as generates detailed question-answer pair(s). Consider for example given an image with two teams of players playing frisbee in ground and enjoying the game shown in Fig~\ref{fig:teaser}. Our proposed method  automatically generate questions that are meaningful, relevant, non-generic and detailed, such as ``\textit{What competitive game a group of guys are playing?}” ,``\textit{How many teams of people are cluttered together during a frisbee game?}”, ``\textit{What is the game in some sport players are competing and having fun?}”. While the questions for the same image in VQA dataset~\cite{1} has generic questions like ``\textit{What sport are they playing?}”, ``\textit{Are all the people in the photo wearing shorts?}”, ``\textit{What numbers are the white players?}” respectively. We see that our proposed VQAG method generates more meaningful and detailed question-answer pairs unlike VQA. In our proposed approach, we first extract answer from the list of objects identified from the image. We use Faster RCNN~\cite{2} object detection technique to extract the objects from the image. Obtaining answers from the extracted objects, we then use these answer to generate meaningful question using associated caption (see Section 4). The proposed VQAG method significantly outperforms interms of question-answer generation compare to SOTA~\cite{15,16}. We firmly believe that our work will boost ongoing research efforts~\cite{1,6,25} in the broader area of conversational AI and scene-text  understanding. Our implementation will be made publicly available on acceptance of the work.

The major contributions of this paper are three folds: 
\begin{itemize}
\item { We draw the attention of the Document Analysis and Recognition community to the problem of visual question answer generation by leveraging image caption  and semantically bridge the visual content with
the associated caption. We are the first to explore the VQAG problem and is an important step in the development of conversational AI and useful in augmenting the training data of image-based question answering.}
\item{ We propose weakly supervised VQA generation method that creates nearest question using caption and visual information, that are then converted to relevant question using in-order traversal by dependency reconstruction method.}
\item{ Exhaustive experimental analysis are performed on proposed VQAG method. Our model significantly outperforms the existing works~\cite{15,16} interms of qualitative and quantitative results. Extensive ablation study on proposed method is investigated.}
\end{itemize}

\section{Related works}

Recently works started studying visual question answering~\cite{1,3,6,17,18,40, newsvideoqa} and they are coming up with various ways to solve edge cases in VQA. However, there are only a few works that focus on generating question-answer pairs for given images ~\cite{8,9, textvqg}. Our goal is to increase the kind of questions that can be generated with correct answers provided they are relevant to the image.
We carry out experiments in generating question-answer pairs in a weakly supervised manner that in turn helps in generating large datasets for solving various VQA tasks.

\subsection{Visual Question Generation (VQG)}
\begin{figure}[h]
  \centering
  \includegraphics[width=\linewidth]{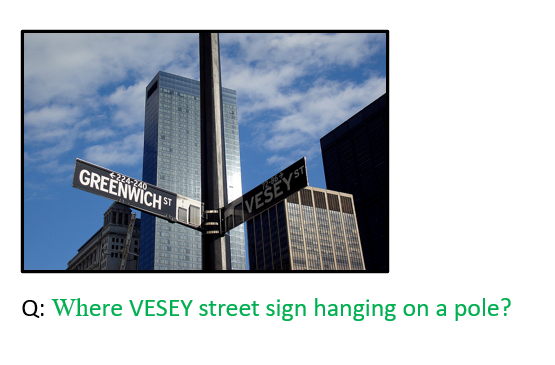}
  \caption{Example question generated by our model which is meaningful, non generic, detailed and relevant to the image.
}
 \label{fig:pole_imvqg}
\end{figure}

VQG~\cite{15, 33, 34} is a well-studied problem in literature and is essential for developing conversational agents and visual dialogue systems~\cite{10,20}. Further, ability to generate relevant and meaningful question by getting in-depth understanding of the visual content of the image is challenging. Few works in this direction are single question generator~\cite{10}, multiple diverse questions generator~\cite{5, 35}, goal-driven question generator~\cite{15, 16}, and learn by context of VQA~\cite{28, 31}. 
In~\cite{15}, the authors generate question based on category of answer. They proposed an information maximizing visual question generator that maximizes the mutual information between image, answer, and answer category during training that helps to generate questions based on the category of answer being asked without answer being given explicitly. But, some of the questions generated in ~\cite{15} based on category of answer does not seem relevant to few images. For example, consider the image shown in Fig~\ref{fig:pole_imvqg}. Given an image of two street signs hanging over a pole with names over it as GREENWICH and VESEY respectively at intersection of roads in front of tall buildings. We aim to automatically generate questions that are meaningful, relevant, non-generic and detailed, such as ``\textit{Where VESEY street sign hanging on a pole?}” but, if given the category as color wrt same image information maximization VQG~\cite{15} might have generated question related to colour of the sky, building or clouds which seems very generic though it is category-based generator. In our method, we overcome this issue by first generating answer and depending on the answer, category-based questions are generated.

In ~\cite{16}, authors generate questions based on answer category but the approach taken was different where the mutual information between image, question, and answer category is maximized at latent space. The variational auto-encoder is used to reduce the level of supervision, but still they depend on manually created datasets for ground truth. In our method, we fill this gap with  effective usage of visual (image) and textual (caption) meaningful information to generate question-answer pair in a weakly supervised manner.

\subsection{Visual Question Answer Generation (VQAG)}

Traditionally, VQAG methods focuses on category-based answer type~\cite{8, 9}.  In ~\cite{8, 9}, authors generate  category-wise synthetic question-answer pairs  from captions using template-based learning. The categorization is based on answer that not only restricts the number of questions, but also the quality of question degrades. However, in our methods, we categorize based on the type of the question which leads to diversity of question-answer pairs that are being generated related to the input image. We consider six types of question words like ``\textit{how many}”, ``\textit{who}”, ``\textit{what}”, ``\textit{which}”, ``\textit{how much}”, and ``\textit{where}". In ~\cite{19}, authors generate question-answer pairs from Wikipedia text and cited documents using context, question, and  answer triplets. Thus, generating question-answer pairs from images is an ongoing research in AI that helps developing agents to connect easily with the scene and communicate with humans.

\section{Proposed approach}

\begin{figure}[h]
  \centering
  \includegraphics[width=\linewidth]{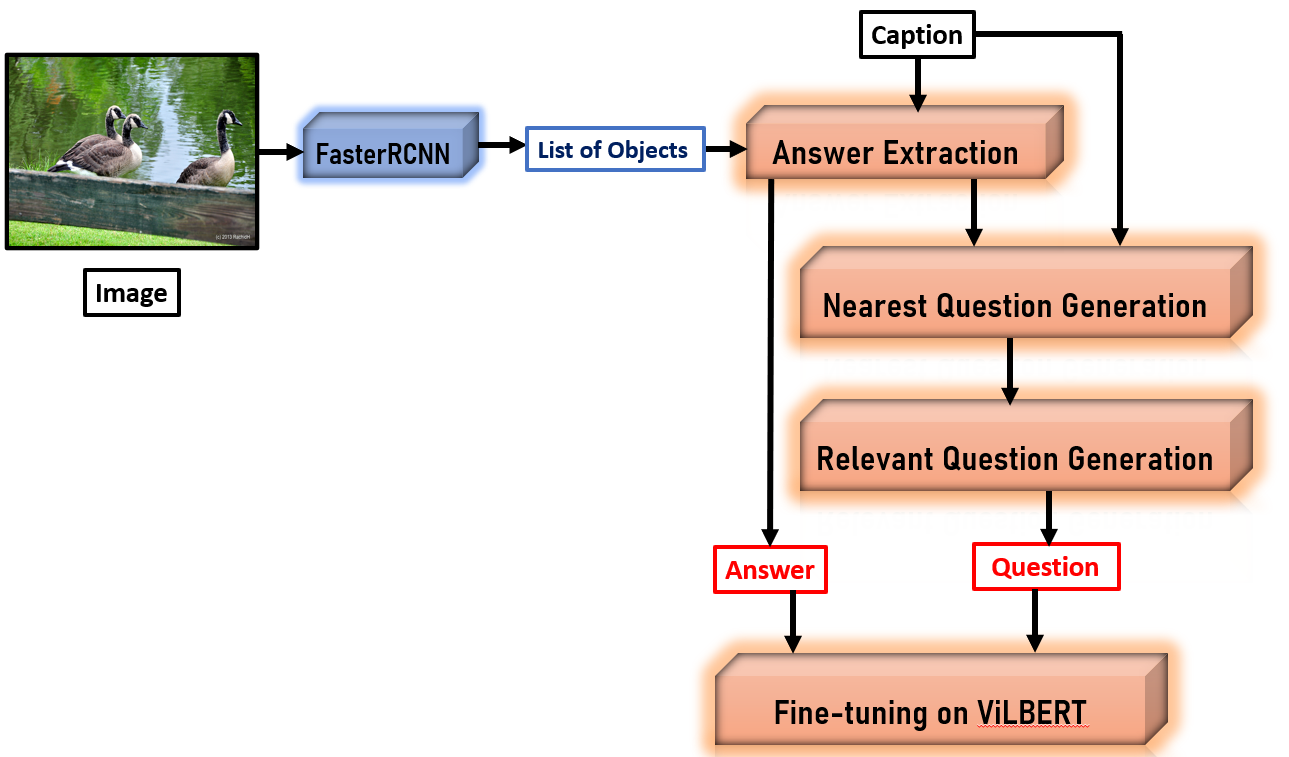}
  \caption{Proposed Visual Question Answer Generator (VQAG) architecture. VQAG has four modules, namely, (i) Answer Extraction, (ii) Nearest Question Generator, (iii) Relevant Question Generator, and (iv) Fine-tuning on ViLBERT.}
  \label{fig:proposed_illustration}
\end{figure}

Given an image and associated caption(s), our aim is to generate a relevant language question such that the answer is in the list of the objects identified from the image. The proposed method should be able to recognize the objects present in the image by properly understanding the visual content, get a clear idea about the caption associated with the image and semantically bridge the textual and visual content inorder to generate meaningful questions. Initially object detection is done using FasterRCNN~\cite{2} followed by a template based method to generate a question from caption and list of objects. After detecting objects, we extract the answer word from caption which is part of the list of detected objects. We, then proceed with question generation. The question generation is a two step process, (i) nearest question generation and (ii)  relevant question generation. The overall architecture of the proposed model is illustrated in Fig~\ref{fig:proposed_illustration}.

\subsection{Answer Extraction}
We first extract the list of objects (Os) from the given input image (I) using pre-trained FasterRCNN~\cite{2} with ResNet-101 as backbone. Since we do not want to lose the information from the image that affects the objects being detected, we set the threshold of confidence to a lower value of $-0.2$. 
Once we detect objects ($O_1,O_2, . . . , O_n$), next   extract the related answer word from the captions ($C_1, C_2, . . . , C_n$). We then check the words from the list of objects identified in last step from the caption. If the word from the list of objects is not found in caption, we then use noun chunkers or name entity recogniser (NER) from Spacy to extract answer words ($W_1, W_2, . . . ,W_n$) from the captions. After extracting the answer words, we mask the answer word in the caption and call it as masked caption.

\begin{algorithmic}
\If{$O_i == <C_1, . . . , C_n>$} 
    \State $ans \gets O_i$
\Else
    \If{$O_i\neq <C_1, . . . , C_n>$}
        \State $ans \gets W_i$
   \EndIf
\EndIf 
\end{algorithmic}

where $W_i$ is answer words from $< W_1,W_2, . . . , W_n>$

\subsection{Question Generation}

\begin{figure}[h]
  \centering
  \includegraphics[width=\linewidth]{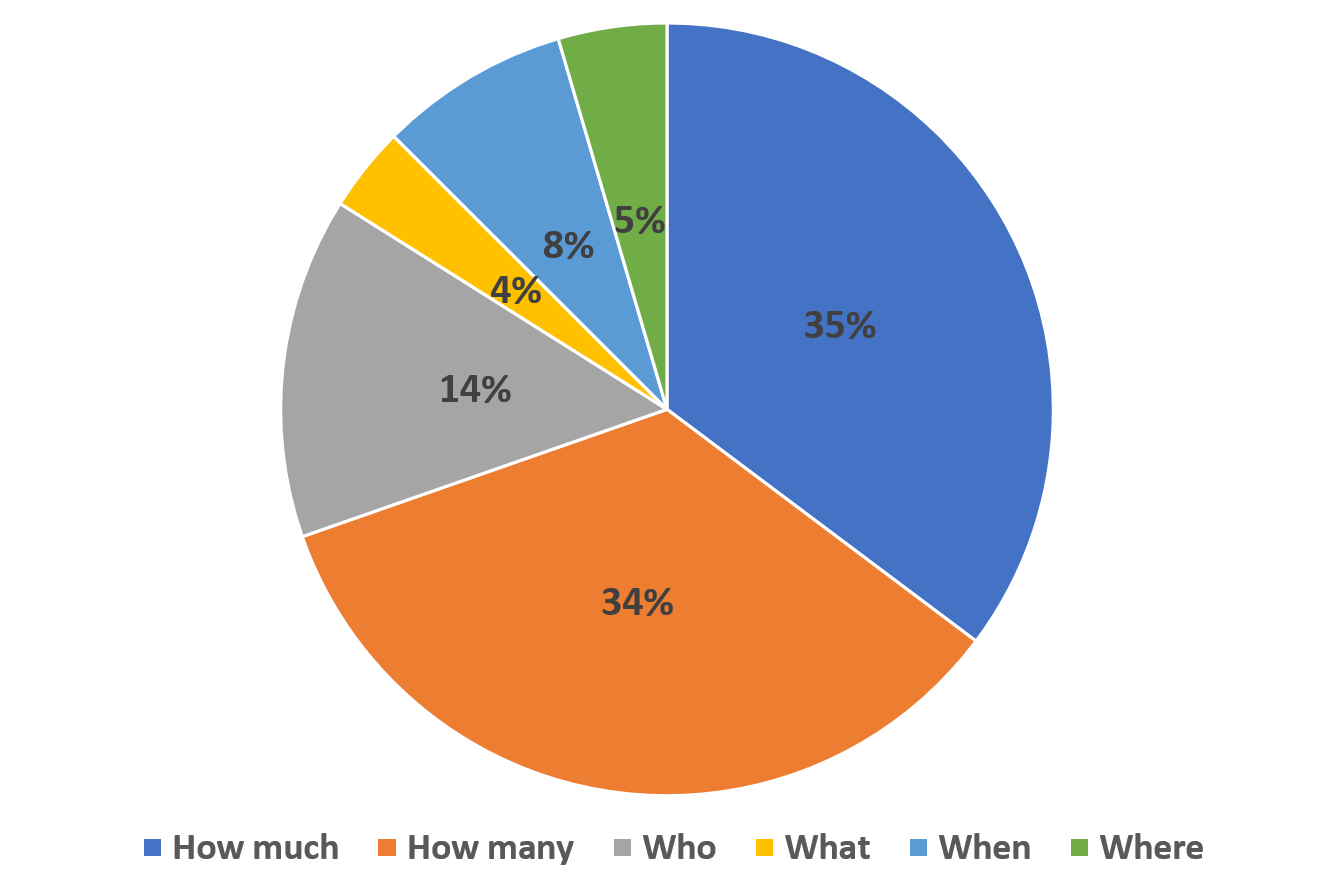}
  \caption{Distribution of different types of question categories generated by our VQAG model.}
\label{fig:distribution}
\end{figure}

We generate nearest questions from the masked caption. We then introduce a rule-based method to rewrite the nearest questions to meaningful relevant questions, which utilizes the dependency structures. Depending on the answer word, Our model generates six types of questions like ``\textit{how many}”, ``\textit{who}”, ``\textit{what}”, ``\textit{which}”, ``\textit{how much}”, and ``\textit{where}". The frequency of occurrence wrt each type of question word is shown in Fig~\ref{fig:distribution}. As question generation is governed by the caption and the detected objects, the distribution of questions are biased towards the count based category type that is ``\textit{how many}” and ``\textit{how much}” in which there is a scope for improvement.

\subsubsection{Nearest Question Generation}

This is the first step in generating questions from the masked caption.
Here, we replace the answer words in the statements with a special mask token, depending on its answer category. Using the masked caption and the answer (with a type label ANIMAL in Fig~\ref{fig:example_method} as masked word is animal in this case), we replace the masked word with special category word which gives us with the Nearest question.

\subsubsection{Translate Nearest to Relevant Questions}
\begin{figure*}[h]
  \centering
  \includegraphics[width=\linewidth]{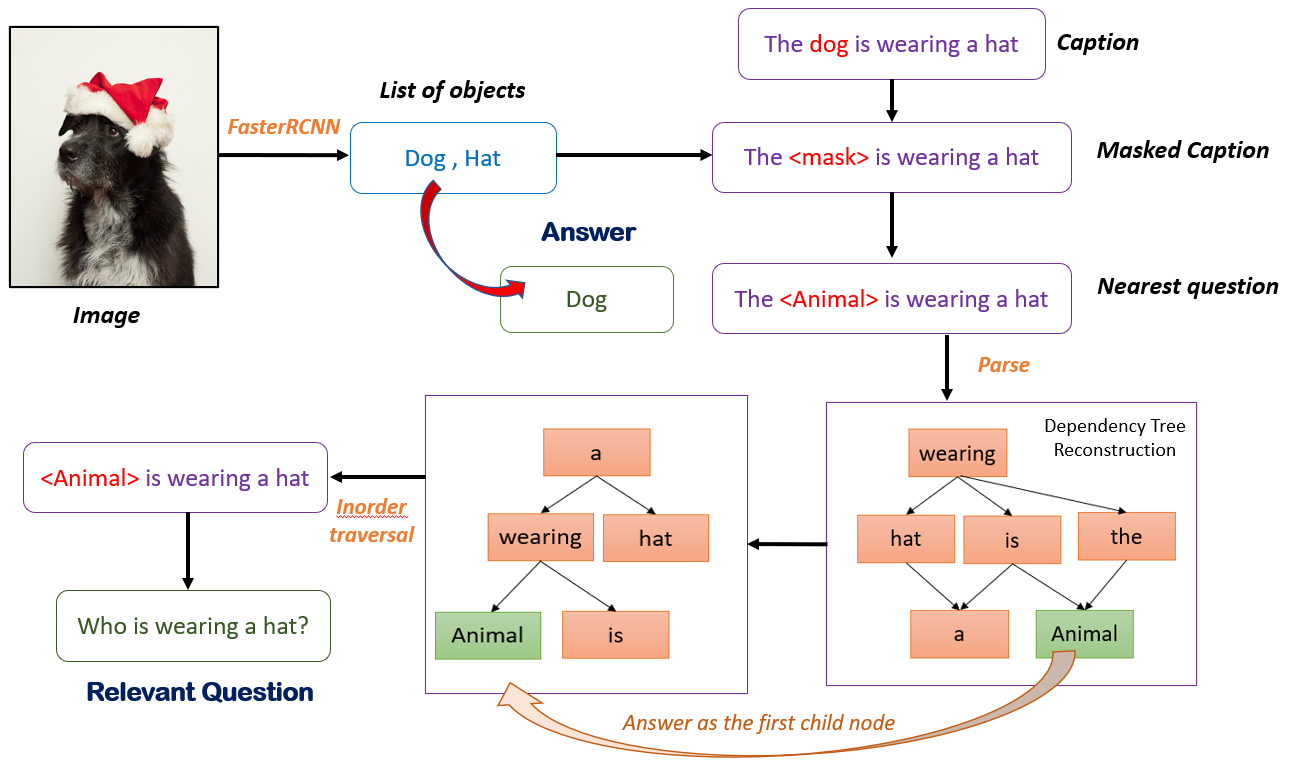}
  \caption{Illustration of the proposed VQAG with an example.}
  \label{fig:example_method}
\end{figure*}

To generate the relevant questions from nearest questions, we perform a dependency reconstruction. First, we create the tree from the words taken from nearest question. Then, we move answer-related words in the dependency tree to the front of the question, as answer-related words are crucial in framing the question. We do this procedure of moving answer related words to the front of the question keeping the intuition that relevant questions usually start with question words.
Dependency parsing is applied to the nearest questions, we follow three steps to translate them to relevant questions, (i)  keep the right child nodes of the answer-related word and prune its lefts, (ii) for each node in the parsing tree, if the sub tree of its child node contains the answer node, we move the child node to the first child node, (iii) finally, do in-order traversal on the reconstructed tree to obtain the relevant question. Using rule-based method mapping, which replaces each answer category with the most appropriate \textit{wh*} word. For example, the LOCATION category is mapped to “\textit{WHERE}” and the COUNT category is replaced by “\textit{HOW MANY}”.
Fig~\ref{fig:example_method} shows the detailed explanation of Visual QA generation following the above mentioned steps.

\subsection{Fine-tuning on ViLBERT}
Generated questions along with the object words that are identified as most appropriate answers are taken as question-answer pairs with their corresponding images and are fine-tuned on popular and state-of-art work in visual question answering~\cite{7}. Based on the  question-answer pairs generated, we create a new vocabulary and then fine-tune on the new vocabulary being appended to the VQA’s vocabulary. After fine-tuning with our questions and answers, we test it on VQA test dataset and obtain a score of $49.2$.The reason for less score is, during the testing time, some of the answer words might be missing that are part of new vocabulary which are not found in ~\cite{1} vocabulary as it is up-streaming task.

\section{Experiments and Results}

In this section, we experimentally validate our proposed method for question-answer pair generation. 
The quantitative and qualitative experimental results of the proposed method and the comparative analysis with SOTA is provided.

\begin{table}
\centering
\caption{\label{tab:table_comparison_m}{Quantitative evaluation of our method against other
weakly supervised SOTA model using standard metrics.}}

\begin{tabular}{c*{7}{c}c}
\toprule

\multicolumn{1}{l|}{ \textbf{Models}} &   \textbf{BLEU}~~~&\textbf{METEOR}~~~&  \textbf{ROUGE-L}\\

\midrule

\multicolumn{1}{l|}{IA2Q~\cite{9} } & 30.42 & 9.42 & - \\

\multicolumn{1}{l|}{V-IA2Q~\cite{9} } & 35.40 & 13.35 & - \\

\multicolumn{1}{l|}{IMVQG~\cite{15} } & 31.2 & 12.11 & 40.27 \\

\multicolumn{1}{l|}{C3VQG~\cite{16}} & 41.87 & 13.60 & 42.34 \\

\midrule



\multicolumn{1}{l|}{Ours} & \textbf{47.78} & \textbf{27.61} & 18.89 \\
\bottomrule

\end{tabular}
\end{table}

\begin{table*}

\centering
\caption{\label{tab:table_comparison_b}{Comparison of our method with various baselines techniques based on context and filtering block.}}

\begin{tabular}{c*{7}{c}c}
\hline

\multicolumn{1}{l|}{ \textbf{Models}} &   \textbf{BLEU}~~~\\

\hline

\multicolumn{1}{l|}{w/o list of objects as context, w/o filtering block} & 45.63
\\

\multicolumn{1}{l|}{w/ list of objects as context, w/ filtering block} & 46.29  \\

\multicolumn{1}{l|}{w/o list of objects as context, w/ filtering block} & 45.13 \\

\hline



\multicolumn{1}{l|}{ w/ list of objects as context, w/o filtering block (Ours)} & \textbf{47.78} \\
\hline

\end{tabular}
\end{table*}

\subsection{Datasets}
Since we consider captions along with images to generate question-answer pair, there is no dedicated dataset available for this task containing question-answer pair with captions for the images. We, therefore, make use of the two popular datasets, namely, MSCOCO~\cite{4} and VQA~\cite{1}. Note that unlike these datasets where originally the task is to generate caption for the image~\cite{19}~\cite{13}and answer a question about the image~\cite{1}~\cite{6} respectively, our aim is to generate question-answer pairs using captions for the images. In other words, given an image and associated  captions (taken from ~\cite{4}), our proposed method learns to automatically generate question-answer pairs similar to the manually curated VQA dataset. 

\subsection{Performance metrics }
We use popular evaluation measures such as bilingual evaluation understudy (BLEU)~\cite{37}, recall-oriented understudy for listing evaluation-longest common sub-sequence (ROUGE-L), and metric for evaluation of translation with explicit ordering (METEOR). The BLEU score compares n-grams of the generated question with the n-grams of the reference question and counts the number of matches. The ROUGE-L metric indicates similarity between two sequences based on the length of the longest common sub-sequence even though the sequences are not contiguous. The METEOR is based on the harmonic mean of unigram precision and recall and is considered a better performance measure in the text generation. Higher values of all these performance measures imply better matching of generated questions with the reference questions.

\subsection{Implementation Details}
We fine-tune our generated QA pair(s) on ~\cite{7} with 4 layer MLP using the Adam optimizer with initial learning rate of 1e−4, batch size of 64. The maximum length of the generated questions is set to 24. We fine-tune the model for 15 epochs. The model is fine-tuned on a single NVIDIA Quadro P5000.

\subsection{Results and discussions }
\subsubsection{Quantitative Analysis}

We evaluate the performance of our proposed method and compare it against three baseline approaches, "without list of objects as context - without filtering block, "without list of objects as context - with filtering block", "with list of objects as context - with filtering block".Explanation for these baselines is provided in section 4.4.2. This comparative result is shown in Table 2 using performance measures discussed in Section 4.2. Note that higher values for all these popularly used performance measure is considered superior. Among the three baseline approaches, "without list of objects as context - with filtering block" generates comparatively better questions. Our proposed final method significantly outperforms all the baseline models. For example our proposed method improves BLEU-score by 1.14 as compared to the most competitive baseline i.e., "without list of objects as context - with filtering block". It should be noted that under these performance measures these gains are considered significant. Further, by design, our proposed method tries to ensure that the answer being object detected in most of the cases and generated question is relevant and meaningful to answer and image.  We also demonstrate our results with and without fine-tuning on ~\cite{7}. It is also seen that fine-tuning on ViLBERT plays a major role in achieving better results which in turn prove that our generated QA pair are meaningful. \\

After fine-tuning on ~\cite{7}, we evaluate our generated questions on VQA questions and our scores are compared with ~\cite{9,15,16}. Our model outperforms all the models as shown in Table 1. The ROUGE-L scores are low compared to the other works as the process of generating questions varies from their approach to our approach. Question generation in their approach follows category based on answers or directly via image features which generates questions similar to that of VQA. Unlike those works we generate questions via captions and detected objects which generates questions way different from VQA.

\subsubsection{Experiments}
We experimented with and without using filtering in our proposed architecture (see Fig~\ref{fig:proposed_illustration}). In case of ``with filtering block” we filter the captions mean we consider the captions only if the particular caption has one of the word from list of detected objects for that image. In this, we filter out certain captions and filtered captions are only considered for further question generation. 
But this approach removed nearly $23$ percent of the captions which in turn lead to lose of good quality questions.

\begin{figure*}[h]
  \centering
  \includegraphics[width=0.85\linewidth]{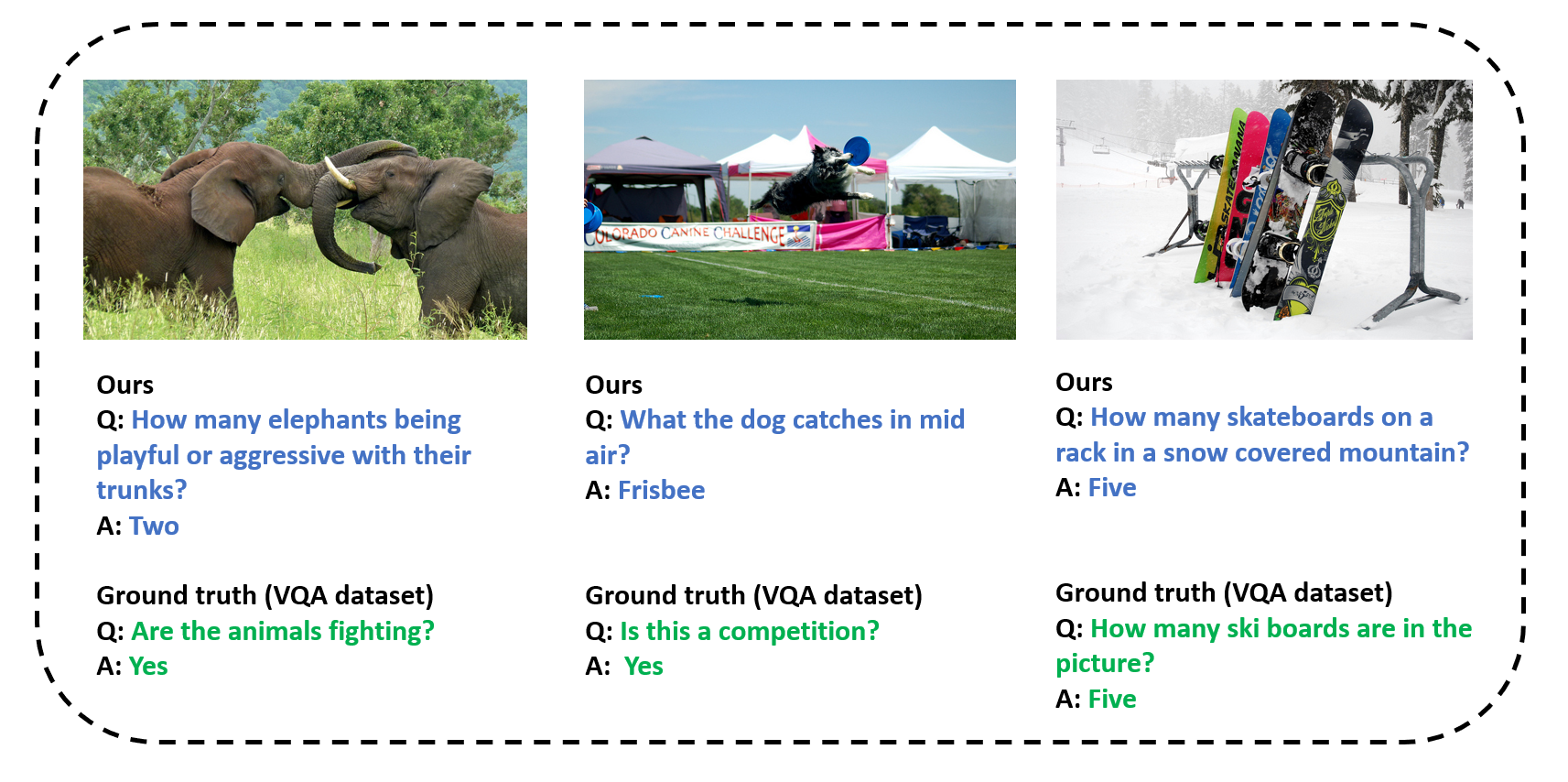}
  \caption{Generated question-answer pairs on VQA dataset~\cite{1}. Results showing from our method and ground truth from VQA dataset respectively. [\textbf{Best viewed in color]}}
    \label{fig:results}
\end{figure*}

\subsubsection{Context as secondary input}

We used context as secondary input to the nearest question generation.
We experimented using “list of objects and caption as context” and only caption as context. Using of context with “list of objects and caption” helped our model to get better questions which are more relevant to the image as it uses list of objects too along with caption.

\subsubsection{Problem with words that are identified as objects}
In one of the baseline models (w/ list of objects as context, w/ filtering block) we had a problem with the certain object words which are identified by object detector (FasterRCNN) that are not found in captions and hence we may lose few good questions there.
We tackled this problem by appending the word identified to the similar words having the same meaning for the detected objects.
For example, we replaced person by adult, man, woman, boy, girl.
However we removed the filtering block as the scores are low compared to the model without filtering block as shown in Table 2. We removed the filtering block from the methodology diagram Fig~\ref{fig:proposed_illustration} as it is not considered in the finalized model.

\begin{figure}[h]
  \centering
  \includegraphics[width=\linewidth]{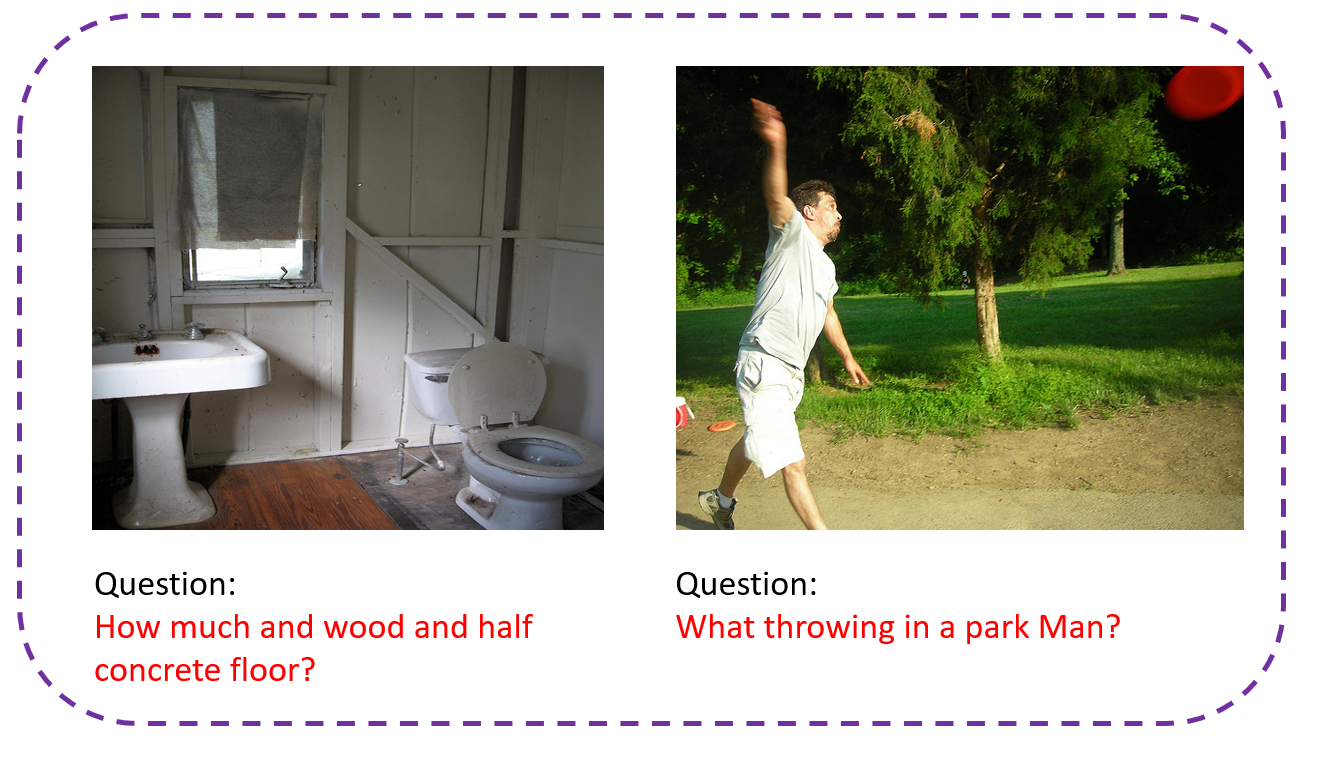}
  \caption{Failure case examples of visual question generation by our model on images from VQA dataset.}
  \label{fig:failure}
\end{figure}

\subsubsection{Qualitative Analysis}

\begin{figure}[h]
  \centering
  \includegraphics[width=\linewidth]{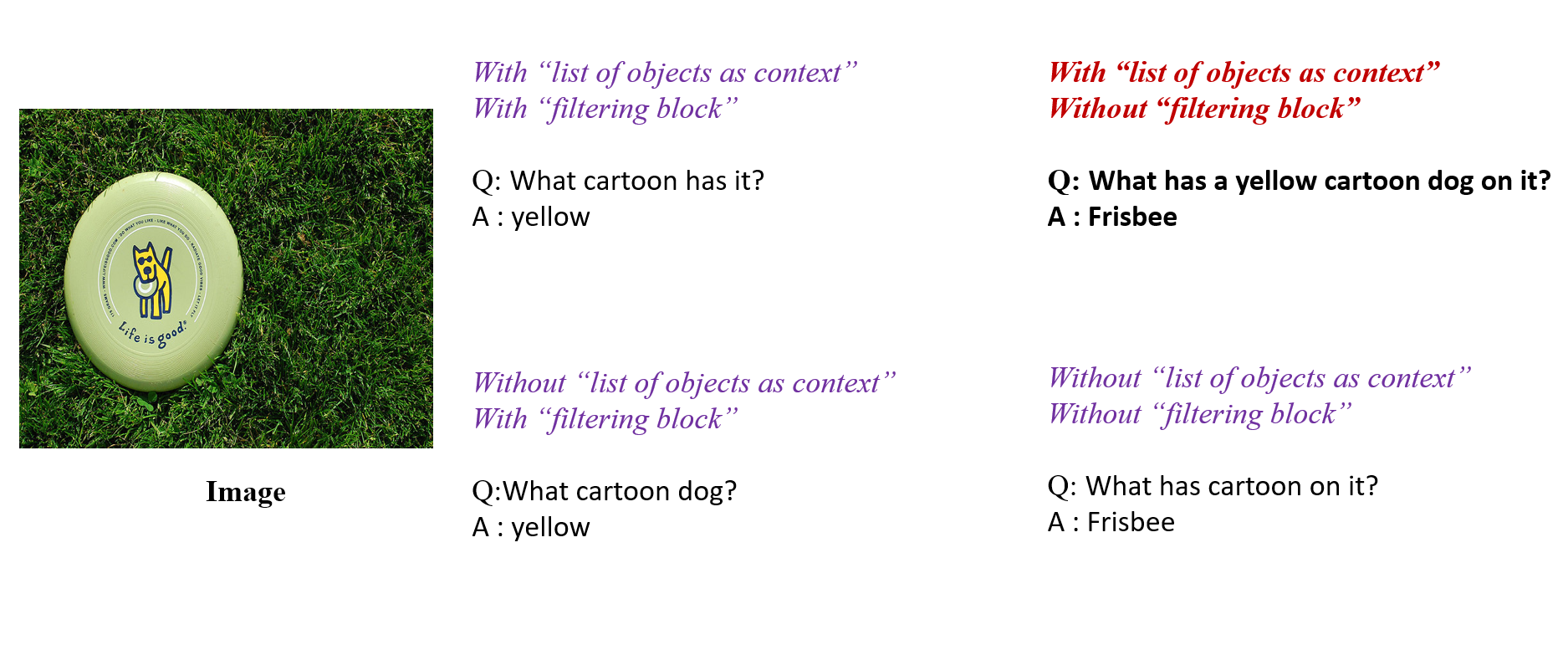}
  \caption{Baseline models and our model qualitative comparison on image from VQA dataset.}
    \label{fig:qual}
\end{figure}

We perform a detailed qualitative analysis of the baselines as well as our proposed method. We first show a comparison of generated QA pair using all the three baselines versus proposed method in Fig~\ref{fig:qual} which is differentiated by color. We observe that the baselines are capable of generating almost linguistically meaningful QA pair with minute chances in sentence formation but not as meaningful as our proposed method. For the given image in Fig~\ref{fig:qual}, the expected question is “ What has a yellow cartoon dog on it?” which is generated by our proposed method. The baseline approaches have generated questions which are appropriate to their corresponding answers but they are not as meaningful as our proposed method. Though the QA pairs generated by baselines are relevant to the input image, as the complexity of the image and the caption increases, the baseline models fail to generate more appropriate and meaningful questions than the proposed model. The proper utilization of visual and textual information generated better questions in our proposed method.

The best baseline model “without list of objects as context - with filtering block” generates question “What cartoon dog?” which is more meaningful than rest baseline approaches but it fails to specify the important information in the image as it is not using list of objects as context and not using filtering block that has removed the caption which dont have one of the words from list of objects, in this case the important information can be Frisbee which is missing. The answer here ( “Yellow” ) is generated by noun chunkers and NER toolkit. Whereas the proposed model generates “What has a yellow cartoon dog on it?” which is more relevant to the image and meaningful than baseline approaches.

Further, more results of our model are shown in Fig~\ref{fig:results} questions and answers generated by our model are in blue, and corresponding ~\cite{1}  questions and answers are in green. Here our model successfully generates meaningful and relevant questions to the image.

The failure of our model pronounced when VQAG misunderstood the scene and output incorrect objects during answer extraction or there is need of generating questions in which objects from image may not act as answer or fails to build the semantic relation between caption and the answer.  
Fig~\ref{fig:qual} shows failure case examples generated by our model. Though the failure case examples does not sense meaningful yet they are relevant to the image.

\section{Conclusions}

We proposed a visual question answer generation method  in weakly supervised manner for a given image and associated caption. Our proposed method properly utilizes the visual properties and generates the question-answer pairs that are meaningful and relevant to the image. Our method has outperformed on SOTA question generating models with a BLEU score value  increased by \textbf{6\%}. Ours is the first work towards developing a visual question-answer pair generation model which considers answer as the one of the object from the image, we restrict our scope to generating questions whose answer is the object from image. Our question-answer pair generator can be used in generating large datasets with no human effort and can also be used in task related to meta-learning and self-supervised learning. Future directions include complex, specific and realistic question-answer pair generation that require deeper semantic reasoning using transformers in understanding image and text together.

\clearpage
{\small
\bibliographystyle{ieee_fullname}
\bibliography{egbib}
}

\end{document}